\RequirePackage{amsthm}

\documentclass[pdflatex,sn-mathphys-num]{sn-jnl}

\usepackage{graphicx}%
\usepackage{multirow}%
\usepackage{amsmath,amssymb,amsfonts}%
\usepackage{amsthm}%
\usepackage{mathrsfs}%
\usepackage[title]{appendix}%
\usepackage{xcolor}%
\usepackage{textcomp}%
\usepackage{manyfoot}%
\usepackage{booktabs}%
\usepackage{algorithm}%
\usepackage{algorithmicx}%
\usepackage{algpseudocode}%
\usepackage{listings}%

\usepackage[shortlabels]{enumitem}
\usepackage{booktabs}
\usepackage{colortbl}

\theoremstyle{thmstyleone}%
\newtheorem{theorem}{Theorem}
\begin{document}

\title[Article Title]{The Role of Fibration Symmetries in Geometric Deep Learning}



\author[1]{\fnm{Osvaldo} \sur{Velarde}}\email{hmakse@ccny.cuny.edu}

\author[1]{\fnm{Lucas C.} \sur{Parra}}

\author[2]{\fnm{Paolo} \sur{Boldi}}

\author*[3]{\fnm{Hern\'an A.} \sur{Makse}}

\affil*[1]{\orgdiv{The Department of Biomedical Engineering}, \orgname{The City College of New York}, \orgaddress{\city{New York}, \state{NY}, \postcode{10031},  \country{USA}}}

\affil[2]{\orgdiv{Computer Science Department}, \orgname{Universit\`a degli Studi di Milano}, \orgaddress{\city{Metropolitan City of Milan}, \postcode{20133}, \state{Milan}, \country{Italy}}}

\affil[3]{\orgdiv{Levich Institute and Physics Department}, \orgname{The City College of New York}, \orgaddress{\city{New York}, \state{NY}, \postcode{10031}, \country{USA}}}

\abstract{
Geometric Deep Learning (GDL) unifies a broad class of machine learning techniques from the perspectives of symmetries, offering a framework for introducing problem-specific inductive biases like Graph Neural Networks (GNNs). However, the current formulation of GDL is limited to global symmetries that are not often found in real-world problems. We propose to relax GDL to allow for local symmetries, specifically \textit{fibration symmetries} in graphs, to leverage regularities of realistic instances.  We show that GNNs apply the inductive bias of fibration symmetries and derive a tighter upper bound for their expressive power. Additionally, by identifying symmetries in networks, we collapse network nodes, thereby increasing their computational efficiency during both inference and training of deep neural networks. The mathematical extension introduced here applies beyond graphs to manifolds, bundles, and grids for the development of models with inductive biases induced by local symmetries that can lead to better generalization.}


\maketitle

\section{Main}\label{sec:intro}

Various disciplines, such as neuroscience, machine learning, and computer science, are interested in finding effective representations for temporal signals, images, grids, or large graphs to solve tasks in reasoning, prediction, or learning \cite{baker_three_2022,poldrack_physics_2021}. One way to find effective representations by identifying \textit{symmetries} in the data. Symmetries are transformations that preserve a particular property of the data \cite{bronstein_geometric_2021}. This idea has been employed for a long time across various research domains, from theoretical physics \cite{noether_invariante_1918} to biological systems \cite{ocklenburg_symmetry_2022,morone_fibration_2020}.

In machine learning, symmetries are crucial for successful neural network architectures \cite{higgins_symmetry-based_2022}. For instance, Convolutional Neural Networks (CNNs) exploit translational symmetry \cite{alzubaidi_review_2021}, useful in image processing where the position of objects can change without changing the appearance of the object. Other examples include Transformers \cite{lin_survey_2021} and Graph Neural Networks (GNNs) \cite{zhou_graph_2020}, which exhibit permutation symmetry, ensuring the output remains unchanged regardless of the order of items in the input. Incorporating symmetries enhances data efficiency and reduces learning complexity \cite{higgins_symmetry-based_2022}. Geometric Deep Learning (GDL) proposes a taxonomy of deep neural network (DNN) architectures based on their symmetries and the symmetry of the data they deal with \cite{bronstein_geometric_2021}. 

Our work focuses on graph symmetries, particularly in graphs with dynamic states, such as those used in GNNs. The predominant symmetry explored in the GDL literature is graph automorphisms \cite{bronstein_geometric_2021, higgins_symmetry-based_2022}, where node permutations preserve the structure of the graph. While useful, preserving global structure is often too restrictive for real-world graphs. We introduce less restrictive levels of symmetries on graphs: \textit{coverings} and \textit{fibration symmetries} \cite{boldi_fibrations_2002}.  These do not preserve the global structure but maintain the graph's dynamic \cite{boldi_fibrations_2002}. Fibration symmetries, in particular, are transformations between directed graphs that preserve the input trees of each node, thus preserving local dynamics. Nodes with isomorphic input trees synchronize \cite{morone_fibration_2020}, meaning they share the same dynamics. For example, in a genetic network, genes that are regulated by the same upstream genes express similarly over time \cite{leifer2021predicting}. Fibrations, being less stringent than automorphisms, appear frequently in biological networks \cite{morone_fibration_2020,leifer2021predicting,avila2024fibration}.

In computational graph networks, symmetries influence both the dynamics of computations and their expressive power. Additionally, fibration symmetries also offer a natural compression method for graph-structured data. For instance, synchronized genes in a gene regulatory network \cite{barbuti_survey_2020} can be treated as a single entity for dynamic analysis, improving the scalability of memory and computations in GNNs \cite{zhou_graph_2020}. Unlike other compression methods \cite{deng_graphzoom_2020,liang_mile_2021}, fibration symmetries ensure the compressed graph preserves the original graph's computations. Previous heuristic compression methods for GNNs, based on neighborhoods \cite{bollen_learning_2023}, show comparable learning performance.

It is important to note that since fibration symmetries preserve the dynamics of the networks without the need to preserve their global structure, this framework can be applied across a wide range of networks in machine learning.


\section{Results}\label{sec:results}

By leveraging \textit{fibration symmetries}, we establish a novel upper bound on the expressive power of GNNs (Section \ref{sec:ExpressivePower}). Additionally, we demonstrate the utility of identifying fibration symmetries within a dataset composed of graphs (Section \ref{sec:Fibrations_data}) for optimizing computations in GNNs while preserving performance (Section \ref{sec:Performance}). Finally, upon observing the emergence of synchronized nodes in DNNs trained via gradient descent (Section \ref{sec:FibrationsMLP}), we propose a new training algorithm that exploits symmetries to optimize DNN architectures (Section \ref{sec:BPfibrations}).


\subsection{Fibration symmetries in GNNs}\label{sec:FibrationGNN}

\subsubsection{Upper bound on the expressive power of GNNs}\label{sec:ExpressivePower}

In this section, we show that the expressive power of GNNs is intimately related to the symmetries of the graph, and extend previous theoretical work to the case of fibration symmetries to obtain tighter bounds on the expressive power of GNNs.

First, we propose a variant of the Weisfeiler-Lehman (WL) test, which we call the \textit{Fibration test}. It takes two graphs $G_1$ and $G_2$ and computes its fibration bases $B_1$ and $B_2$. If the bases $B_1$ and $B_2$ are different, the test will say that $G_1$ and $G_2$ are not isomorphic. Both the WL test and Fibration test are necessary conditions to indicate the existence of isomorphism between graphs; however, the advantage of the Fibration test is that we have reduced the number of operations, as only the input trees need to be computed. On the other hand, we show the operations in GNNs depend exclusively on the in-neighbors of the nodes (see Section \ref{sec:Fibrations_GNNs} and SI). Therefore, for the Fibration test and GNNs, the relevant structure is only in the input trees, not the output trees as considered by the classical WL test.

\begin{figure}[ht]
\centering
\includegraphics[scale=0.25]{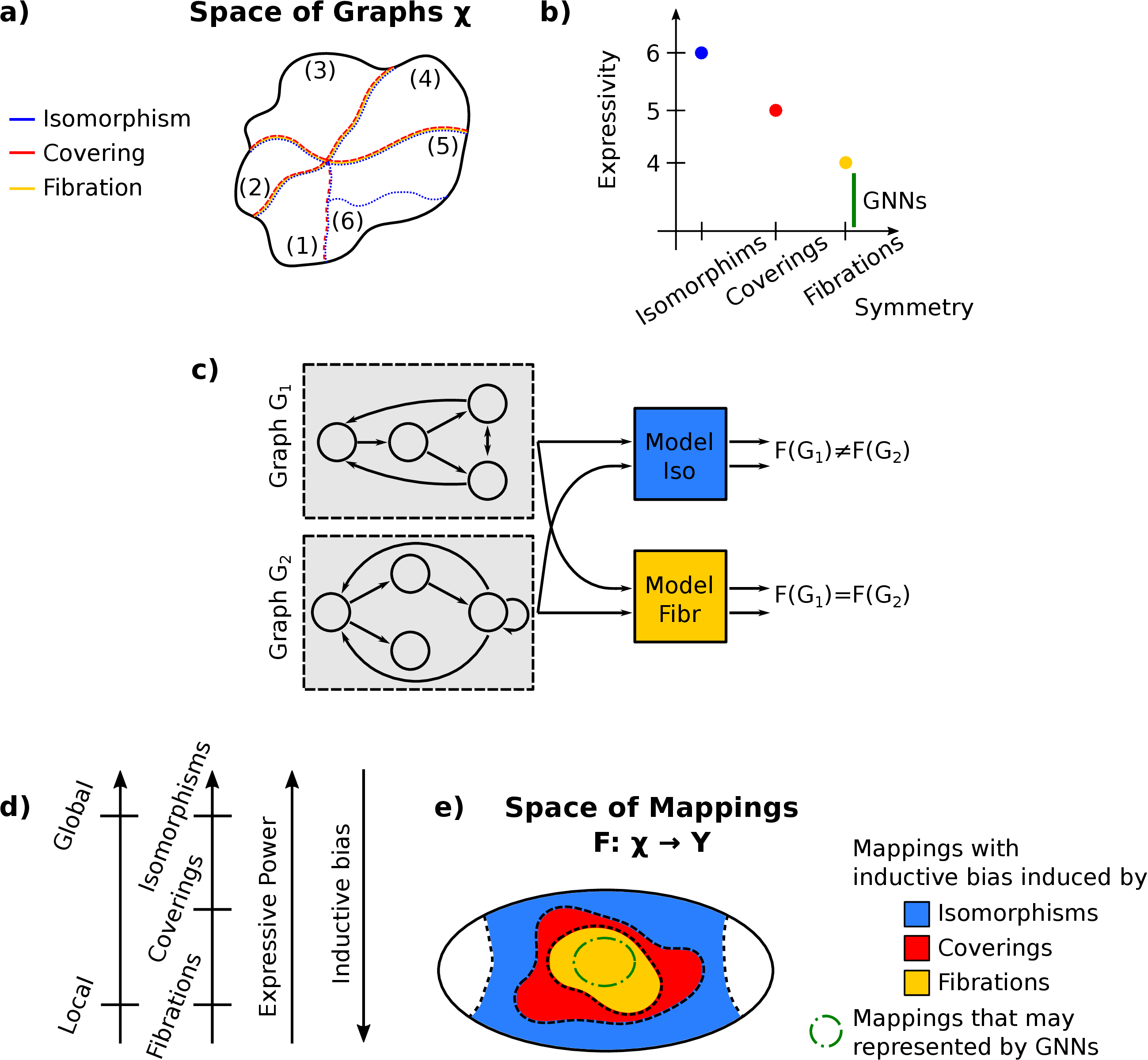}
\caption{\textbf{Symmetries, inductive bias and expressive power}. \textbf{a)} For each level of symmetries (i.e., fibrations, coverings, automorphisms), there is a partition of space of graphs $\mathcal{X}$ (see colored dashed lines). The partition induced by isomorphisms (e.g., 6 classes) is finer than the partition induced by coverings (e.g., 5 classes); and the latter is finer than that induced by the fibrations (e.g., 4 classes). \textbf{b)} Functions with inductive bias induced by isomorphisms have greater expressive power than those that satisfy the inductive bias induced by local isomorphisms (i.e., coverings) or induced by local in-isomorphisms (i.e. fibrations). GNNs are models with inductive bias induced by fibrations and the upper bound on its expressive power is determined by Fibration Test (see Lemma 1). \textbf{c)} The graphs $G_1$ and $G_2$ are not isomorphic but have the same fibration base. A map $F$ with inductive bias induced by isomorphisms (blue box) returns different outputs for the graphs ((i.e., $F(G_1) \neq F(G_2)$). On the other hand, a map $F$ with inductive bias induced by fibrations (yellow box) returns the same output for both graphs (i.e., $F(G_1)=F(G_2)$). \textbf{d)} Isomorphisms are global symmetries, while coverings and fibrations are local symmetries. Isomorphisms, coverings, and fibrations induce different inductive biases, which increase when the symmetry is weaker. This implies that expressive power is lesser for models that use weaker symmetries. \textbf{e)} In the Space of Mappings, there is a subspace of maps that satisfies the inductive bias induced by isomorphisms (blue area), coverings (red area), and fibrations (yellow area). GNNs are a subset of the maps with bias induced by fibrations (see green line).}
\label{fig:summary}
\end{figure}

Based on the preceding comments, we can reformulate and extend the theorems of \cite{xu_how_2019} in terms of fibration symmetries and Fibration test:

\begin{theorem}
Let $G_1$ and $G_2$ be graphs with minimal fibrations
$\phi_i: G_i \rightarrow B_i$ ($i=1,2$) and $B_1 \not\sim B_2$. If a
graph neural network $\mathcal{A}: \mathcal{X} \rightarrow
\mathbb{R}^d$ maps $G_1$ and $G_2$ to different representations, then
the Fibration test also decides that $G_1$ and $G_2$ have
different minimal bases.
\label{lemma}    
\end{theorem}

\begin{theorem}
Let $\mathcal{A}: \mathcal{X} \rightarrow \mathbb{R}^d$ be a GNN. With
a sufficient number of GNN layers, $\mathcal{A}$ maps any graphs $G_1$
and $G_2$ that the Fibration test decides that the graphs
have different minimal bases, to different representations if the
following conditions hold:
\begin{enumerate}
    \item $\mathcal{A}$ aggregates and updates node features iteratively with 
    \begin{equation*}
        h_v^{(k)} = \gamma \left( h_v^{(k-1)}, f \left( \{\{ h_u^{(k-1)}, u \in N(-,v) \}\}) \right)\right)
    \end{equation*}
    where the functions $f$ and $\gamma$ are injective.
    \item $\mathcal{A}$'s graph-level readout is injective.
\end{enumerate}
\label{theorem}
\end{theorem}

These theorems demonstrate that an upper bound on the expressive power of GNNs is the expressive power of the Fibration Test, which is lower than the expressive power of the WL test. We emphasize that for undirected graphs, input trees and output trees are the same; therefore our results are identical to those of Xu {\it et al.} \cite{xu_how_2019}.

For each level of symmetry (i.e., isomorphisms, coverings, fibrations), we define an inductive bias (Section \ref{sec:InductiveBias}) and a partition of the Space of Graphs (Fig. \ref{fig:summary}a). Intuitively, there are more graphs with the same fibration base than isomorphic graphs. Thus, there are more equivalence classes for stronger symmetries. For each inductive bias, we calculate the maximum expressive power and find the order $E_{F_{fib}} \leq E_{F_{cov}} \leq E_{F_{iso}}$ (Fig. \ref{fig:summary}b). Intuitively, each inductive bias incorporates information about the symmetry utilized: the map $F$ does not distinguish between graphs in the same class (Fig. \ref{fig:summary}c). In Fig. \ref{fig:summary}d, we show how the expressive power and the inductive bias vary with the symmetries. Global symmetries (e.g. isomorphisms/automorphisms) are stronger symmetries that induce lesser inductive biases and bigger expressive powers. For instance, the strictest equivalence relation is graph equality; in this case, any $F$ satisfies the bias condition, i.e. there is no bias applied to the mapping at all. However, the least strict equivalence relation is that all graphs are equivalent; in this scenario, the mapping $F$ must be constant, i.e. a very strong inductive bias. 

Another way to visualize the expressive power is through the volume of the set of mappings that satisfy a certain bias condition (see Fig. \ref{fig:summary}e). The set of mappings with inductive bias induced by fibrations (see yellow area) is a subset of those with inductive bias induced by coverings (see red area). This latter set is also a subset of the functions with inductive bias induced by automorphisms (see blue area). GNNs are particular functions that satisfy the bias induced by fibrations (see green line inside the yellow area), as they depend solely on the in-neighborhoods.

These statements are about the expressive power of a GNN. A different and equally important question is about the performance of a GNN, i.e. how well does a model learn to perform a particular input-output mapping for a real-world dataset? 

\subsubsection{Fibration symmetries in graph-structured datasets}\label{sec:Fibrations_data}

\begin{figure}[ht]
\centering
\includegraphics[scale=0.19]{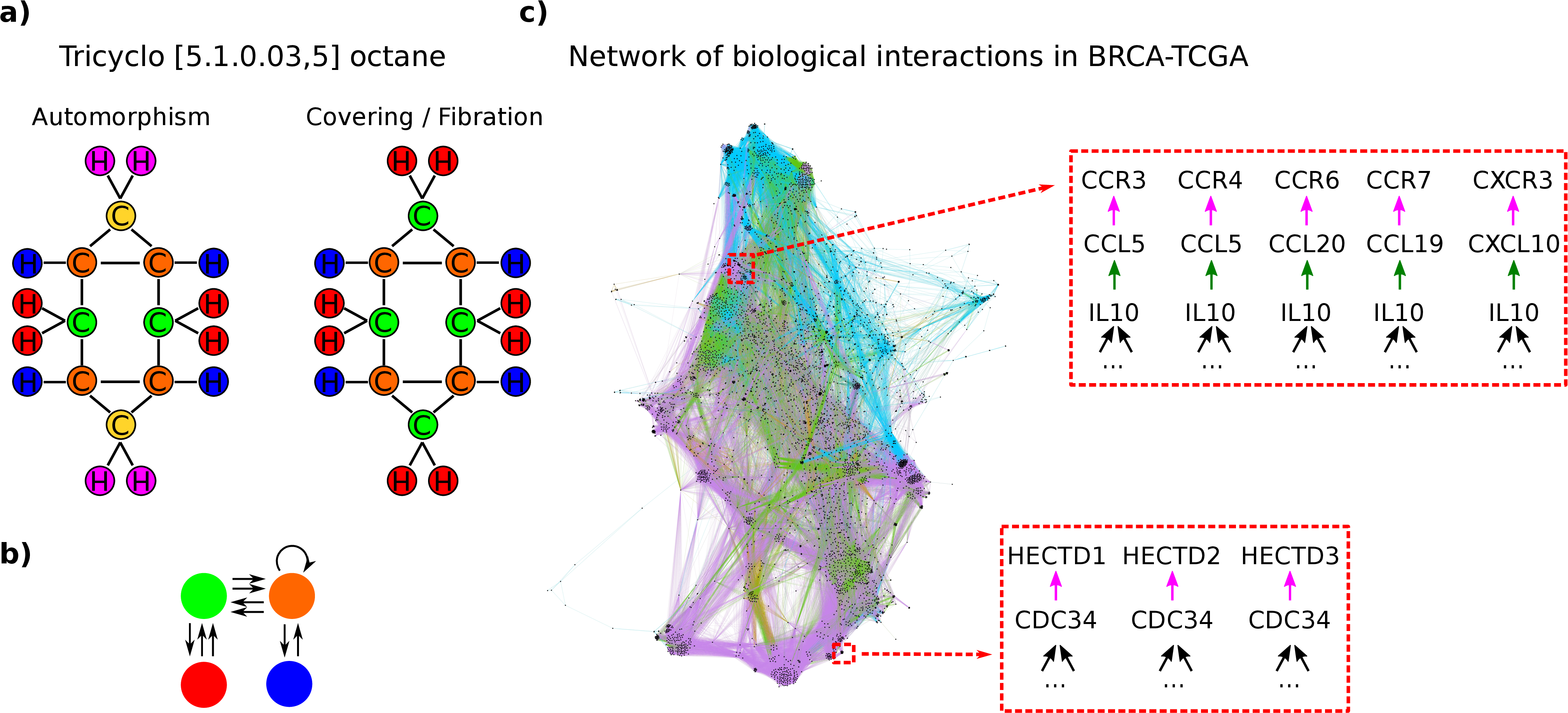}
\caption{\textbf{Fibration symmetries in datasets}. Examples of the fibration symmetries observed in the graphs of QM9 and BRCA-TCGA datasets. \textbf{a)} 
Tricyclo [5.1.0.03,5] octane is a molecule composed of eight carbon atoms and twelve hydrogen atoms. The partition induced by automorphisms consists of six orbits, while the partition induced by fibrations consists of four fibers. One fiber consists of 4 hydrogen atoms (blue circles), one fiber of 8 hydrogen atoms (red circles), and two fibers of 4 carbon atoms (green and orange circles) \textbf{b)} Fibration base of Tricyclo octane. \textbf{c)} The biological interaction network in BRCA-TCGA consists of 9288 genes interacting in various ways represented by arrows of different colors. The genes CCR3, CCR4, CCR6, CCR7, and CXCR3 are in the same fiber because they have isomorphic input trees. The genes CCL5, CCL20, CCL19, and CXCL10 (resp. HECTD1, HECTD2, and HECTD3) are in the same fiber because they receive information from the same gene IL10 (resp., CDC34).}
\label{fig:fibrations_datasets}
\end{figure}

Here we report fibration symmetries and the corresponding base graph in two real-world datasets
(Fig. \ref{fig:fibrations_datasets}), namely, the Quantum Mechanic 9 (QM9) dataset that provides quantum chemical properties for a large number of small organic molecules; and The Cancer Genome Atlas Breast Cancer (TCGA-BRCA) dataset that includes gene expression and survival
time of individual patients with breast cancer (see descriptions in Section \ref{sec:datasets}). Both datasets consist of graph-structured samples. We calculate the minimal fibration base $B$ of each graph $G$ using the ``minimal balanced coloring"
algorithm (see Section \ref{sec:met_fibration}).  

Some examples of the graphs involved and their corresponding partitions are shown in Fig. \ref{fig:fibrations_datasets}. From the QM9 dataset, we show the structure of Tricyclo [5.1.0.03,5] octane. The graph structure of Cyclopropylmethanol in Fig. \ref{fig:fibrations_datasets}a has 4 fibers (see panel Coverings/Fibrations). To maintain the identity of the atoms within a fiber, there cannot be two different types of atoms. The partition induced by the fibration and covering symmetries coincide since the graph is undirected. However, they are not necessarily the same as the partition induced by automorphisms. For this molecule (Fig. \ref{fig:fibrations_datasets}a), there are six orbits derived from the horizontal and vertical reflection symmetries (see panel Automorphisms). For each molecule, we build the fibration base; e.g., in Fig. \ref{fig:fibrations_datasets}b, we show the base of Tricyclo octane. Note that the compression factor (i.e., the number of nodes in the base divided by the number of nodes in the molecule) for Tricyclo octane is 4/20 = 0.2. This value depends on the molecular structure. We found that the average compression factor across all samples in the QM9 dataset is $\sim 74\%$.

In both datasets, there are structures with different types of interactions. For example, molecules have simple, double, and/or triple bonds. In the biological interaction network of the TCGA-BRCA dataset, some interactions represent the expression control of a gene (green arrows), activation/inhibition of protein complex components (pink arrows), or reaction control of a protein (light blue arrows). This dataset includes only one graph of 9288 genes with a compression factor of $\sim 53\%$. In Fig. \ref{fig:fibrations_datasets}c, we show examples of fibers in the network. 

\subsubsection{Training of GNNs with minimal fibration base maintains performance}\label{sec:Performance}

We implement two GNNs (see details of the models in \hyperlink{https://www.github.com/MakseLab}{GitHub}) to learn the following prediction tasks from the data:
\begin{enumerate}[start=1,label={(\bfseries T\arabic*):}]
\item Prediction of the dipole moment of each organic molecule based on their structure.
\item Prediction of the overall survival time of each patient based on their gene expression profile. 
\end{enumerate}

\begin{figure}[ht]
\centering
\includegraphics[scale=0.14]{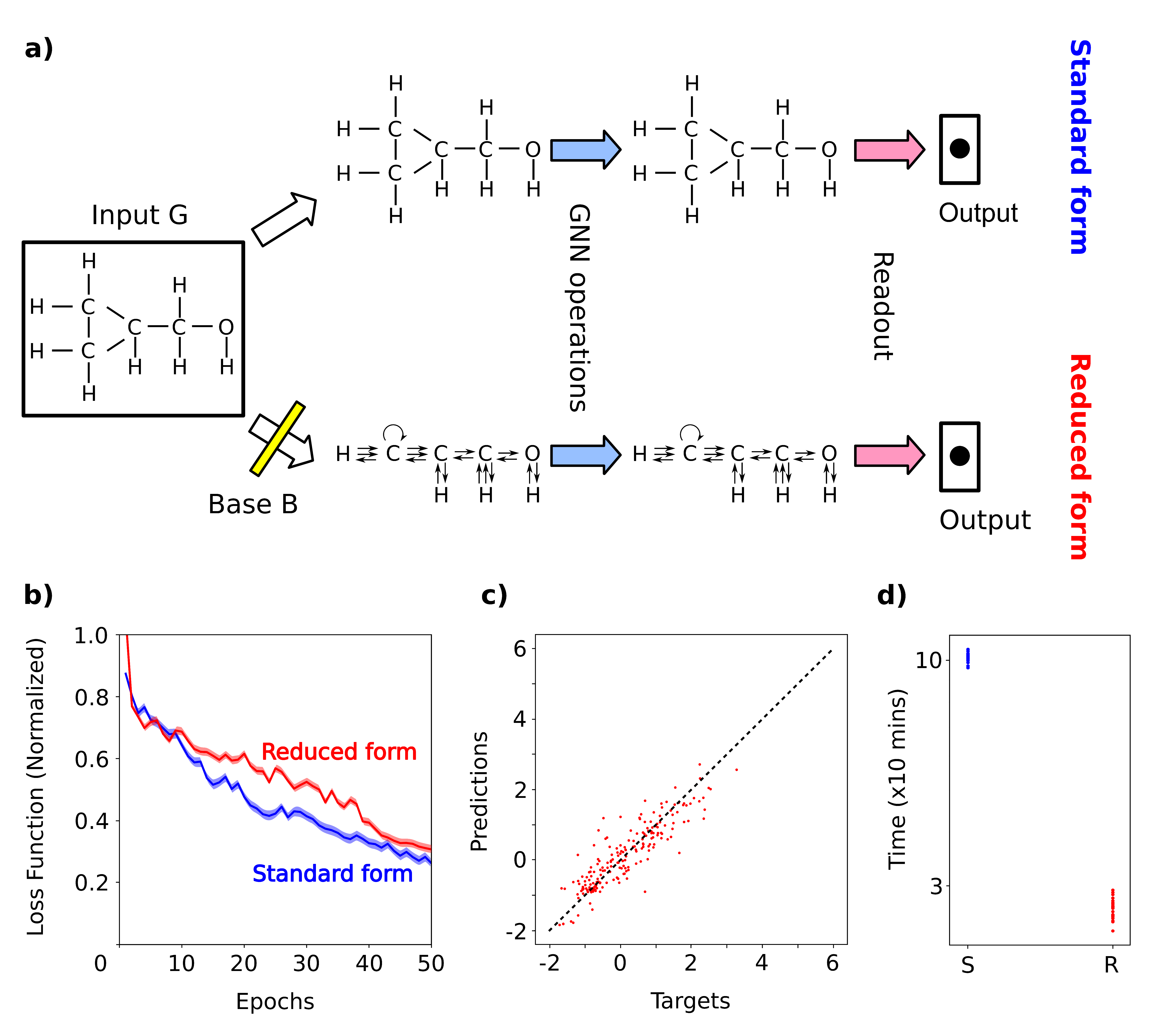}
\caption{\textbf{Standard and Reduced form of GNNs}. a) The structure of a GNN and the number of operations are derived from the structure of the input graphs $G$ (Standard form). Finding the fibration base $B$ for the input graph $G$ allows us to reduce the number of operations in GNNs (Reduced form). b) Loss function during the training stage for the task T1. c) Predictions of the network vs. dipole moment of the molecules. d) Comparison of the execution times for ``standard form" and ``reduced form" for task T1.}
\label{fig:performance}
\end{figure}

We train the networks for \textbf{T1} and \textbf{T2} using QM9 and BRCA-TCGA datasets, respectively. The training was performed both in the standard and reduced forms (see Fig. \ref{fig:performance}a) as explained in Section \ref{sec:Fibrations_GNNs}. In Fig. \ref{fig:performance}b-d, we present the results for \textbf{T1}. 

We observe that the evolution of the loss function over the training steps for both forms is similar (see Fig. \ref{fig:performance}b). In particular, after 50 epochs, the difference is only $4\%$. In Fig. \ref{fig:performance}c, we show the predictions of the GNNs versus the actual dipole moment values for the molecules in the test set. The loss function is calculated as the mean squared error of these values (i.e., distance to the diagonal). A relevant advantage of the reduced form is its execution time, which is 4 times shorter than the time for the standard form (see Fig. \ref{fig:performance}d). It is important to note that the reduction in the execution time is much greater than the reduction in the samples. We do not show the results for \textbf{T2}; however, these can be generated using the code available on GitHub. The results are consistent with the behaviors observed in Fig. \ref{fig:performance}.

Given that GNNs satisfy bias induced by fibrations, it is expected that the learning process remains invariant under the fibration symmetry (i.e. graph compression from $G$ to $B$). Our results confirm empirically that the performance of a GNN is equivalent when trained on either the original or compressed graphs of real-world datasets. In other words, we leveraged that every (learning) procedure executed on the base replicates the same procedure executed on the original graph network. This approach improves computational efficiency without compromising learning performance. 


\subsection{Fibration symmetries in MLPs}

\subsubsection{Synchronization of nodes during the training of MLPs}\label{sec:FibrationsMLP}

\begin{figure}[ht]
\centering
\includegraphics[scale=0.25]{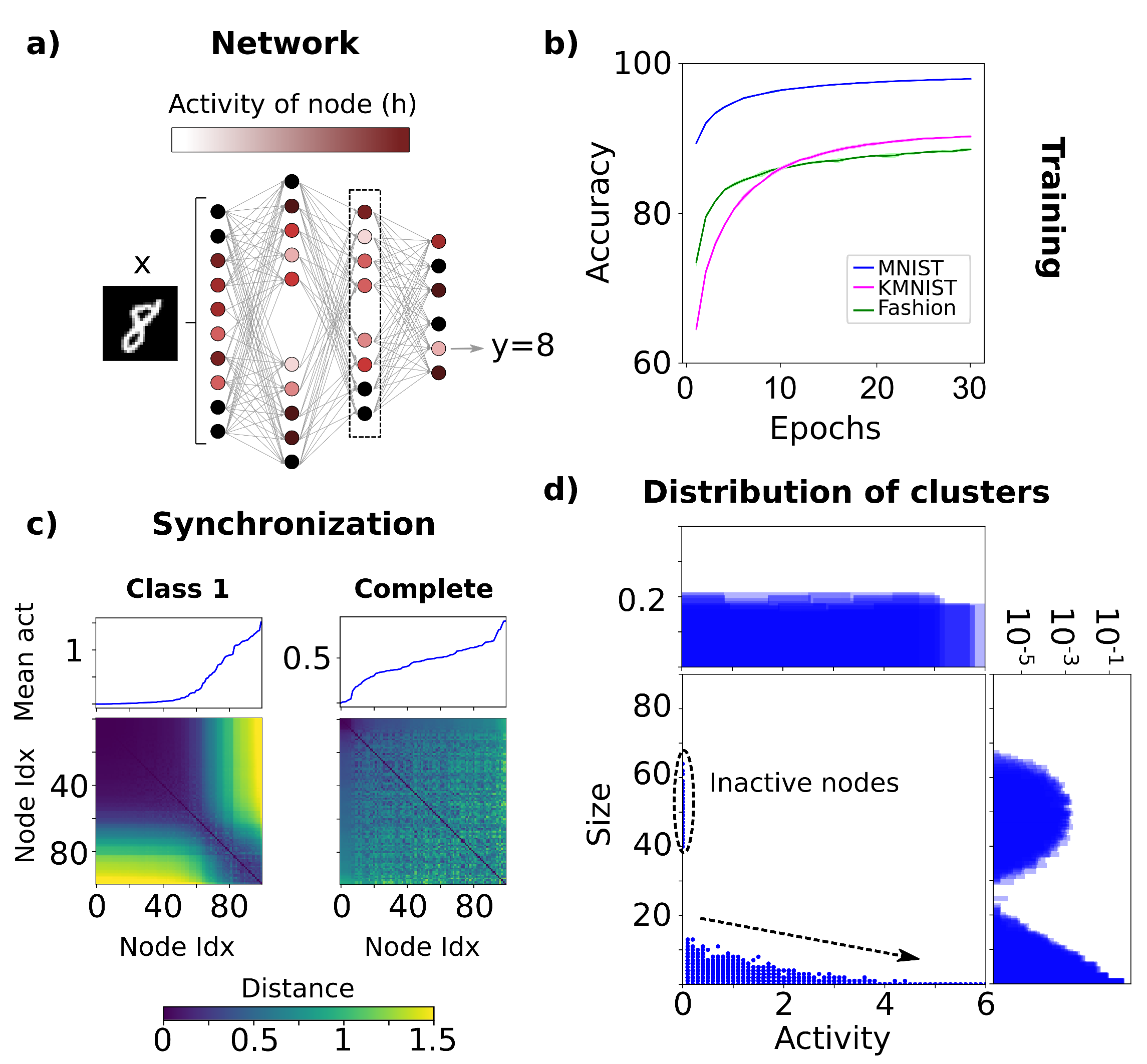}
\caption{\textbf{Synchronicity in MLP}. \textbf{a)} An MLP trained to classify images (e.g. digits in MNIST dataset). \textbf{b)} Evolution of MLP accuracy during the training stage for different datasets: MNIST, KMNIST, and Fashion. \textbf{c)} After the training stage, clusters of synchronized nodes are detected. The clusters depend on the data used $\mathcal{S}$. The results presented in the left (resp., right) column correspond to cases where $\mathcal{S}$ is the subset of samples from class 1 (resp., from all classes) within the test set. The distance matrix between nodes $\Lambda$ (see definition in Methods) is represented by a color map. The node indices are sorted by mean activity across all the samples in $\mathcal{S}$ (see Mean act). \textbf{d)} In the Size vs. Activity plane, we show the distribution of cluster size and activity across input images (i.e., $\mathcal{S}$ is a single image). In the upper (resp., right) panel, the marginal distribution of cluster activities (resp., sizes) is plotted.} 
\label{fig:MLP}
\end{figure}

Here we ask whether fibration symmetries also play a role in conventional DNNs, such as the MLP: Multi-Layer Perceptron (Fig. \ref{fig:MLP}a) which has been taken as a prototype for neural networks in the brain. A common observation in the brain is oscillatory activity, where the activity of neurons changes in unison, i.e. they are synchronized \cite{thut_functional_2012}. 

When first initialized, MLPs have a random weight on the graph structure giving rise to diverse ``activation" in each neuron. We hypothesized that during learning the weights adjust to give synchronized network activity, i.e. two or more nodes converge during learning to have the same ``activation" values for all input samples. To test this, we analyze the activity of the nodes in MLPs that were trained and achieve accuracy of 97.67\%, 90.58\%, and 88.42\% (Fig. \ref{fig:MLP}b) for MNIST, KMNIST, and Fashion datasets (see Section \ref{sec:datasets}), respectively.

In Fig. \ref{fig:MLP}c, we plot the mean activity of nodes in the second hidden layer after the
training process (see panel ``Synchronization"). The mean activity is calculated as the average of each node's activity across different samples. The node indices are sorted by mean
activity. When samples are from the same class (e.g., samples of class 1 in MNIST), we observe large, well-defined clusters of synchronized nodes compared to when synchronization is determined across the entire dataset (see color map in Fig. \ref{fig:MLP}c). Moreover, at the beginning of training, the activity values $h_i$ are of the order of $\epsilon_l$, suggesting the presence of only one trivial cluster in the distance matrix (data not shown).

Similar to previous cases, clusters (or fibers) of different sizes and associated activity values appear for each sample. In other words, we obtain a distribution of fibers across the samples of the dataset. In Fig. \ref{fig:MLP}d, we show the distributions of the size and activity of the fibers across input images. Note that the marginal activity distribution is uniform, while the marginal size distribution is bimodal. One mode is associated with fibers of inactive nodes ($h_i = 0$) whose typical size is around 40-60\% of the total nodes. The zero activity of these nodes arises from the ReLU activation function applied to non-positive inputs. The rest of the nodes cluster such that clusters with higher activity are smaller. 

\subsubsection{Application of fibration symmetries in gradient descent}\label{sec:BPfibrations}

\begin{figure}[ht]
\centering
\includegraphics[scale=0.3]{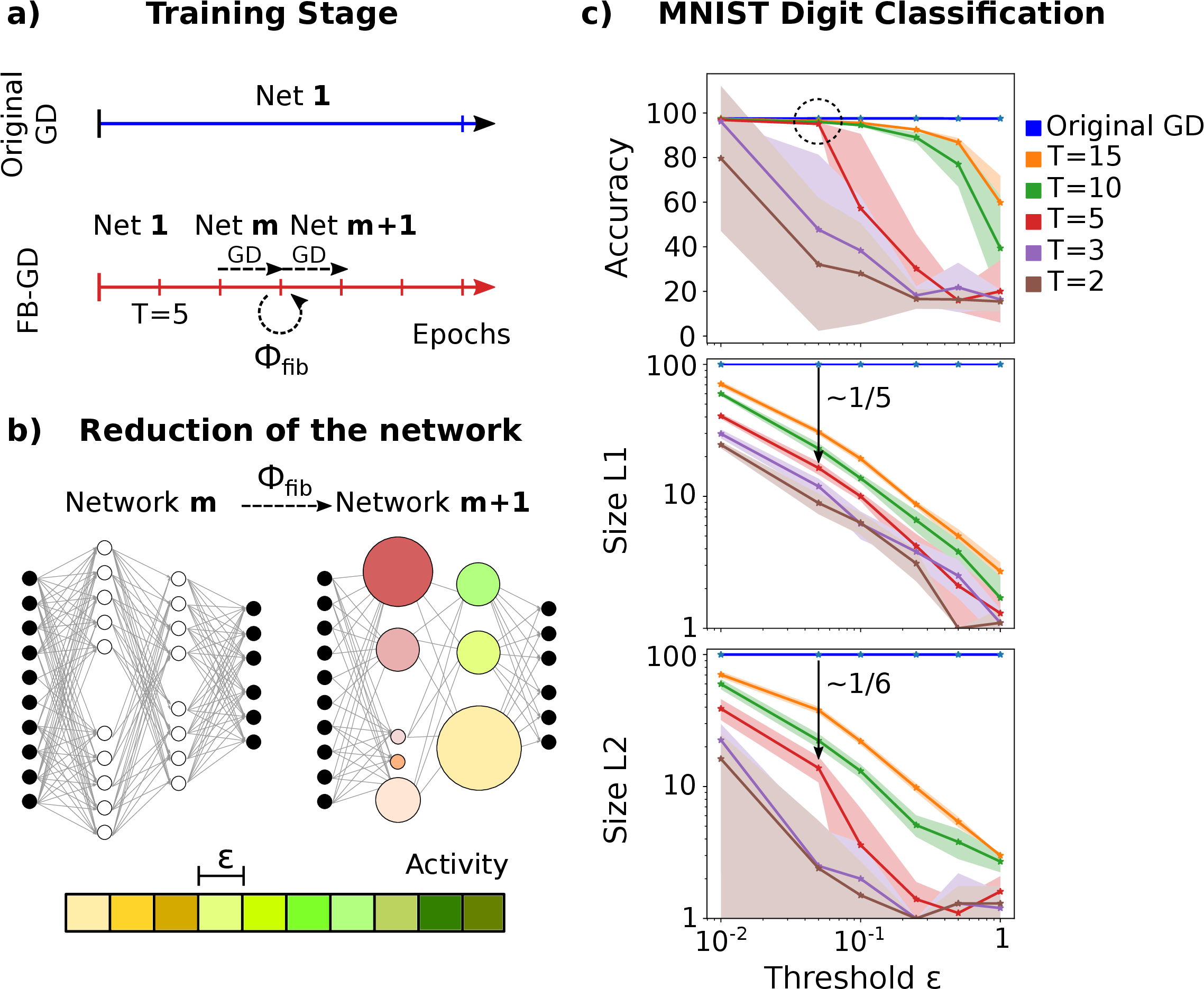}
\caption{\textbf{Fibration Gradient Descent (FB-GD)}. \textbf{a)}
  Usually, during the training stage of an MLP via gradient descent,
  the network's architecture remains unchanged over time (see Net 1 -
  Original DG). For FB-GD, both the connection weights and the network structure are updated (see FD-GB). Every $T$ epochs, clusters of nodes are detected to collapse them into a single node, similarly to constructing the bases (Net m+1) of a graph (Net m) with fibration symmetries $\Phi_{fib}$.  \textbf{b)} Each
  cluster (colored circles) is associated with an activity interval of size $\epsilon$ (see color bar). The update equation for the
  parameters of the Network m+1 is designed to be compatible with the update equation for the previous Network m. \textbf{c)} The final size of the network (see Size L1/L2) and its performance (see Accuracy) depend on $\epsilon$ and $T$. In the limit $\epsilon \rightarrow 0$ and $T \rightarrow \infty$, the results of FB-GD coincide with those of the original gradient descent.}
\label{fig:results_bp}
\end{figure}

We apply the node synchronization property of the fibration symmetries to modify the gradient descent method, which is used to minimize the loss function during the training stage. More precisely,
with this new version \textit{Fibration-Gradient Descent} (\textit{FB-GradDesc}), we construct a smaller network $B$ (i.e.,
base) composed of fibers from the original network $G$ (see
Fig. \ref{fig:results_bp}a-b), and update the parameters of $B$ using
the information from the parameters of $G$. The way we update the
parameters ensures that the input information of each node is
preserved, as required by the fibration symmetries. The
deduction and implementation of FB-GradDesc are explained in
Section \ref{sec:methods_mlp}.

The proposed definition of synchronicity (Section \ref{sec:met_fibration_mlp}) depends on the set of samples used $\mathcal{S}$ and the threshold $\epsilon$ (see color bar in Fig. \ref{fig:results_bp}b). For FB-GradDesc, we use $\mathcal{S}=\mathcal{D}_{\text{training}}$ and determine the optimal threshold $\epsilon$. Additionally, we use a time constant $T$, which represents the number of epochs required to detect the clusters and
construct the base $B$. In other words, for every $T$ epochs,
FB-GradDesc calculates the fibers using the synchronicity
criterion and reduces the size of $G$.

The definition of synchronicity becomes stronger and more stringent as
$\epsilon \rightarrow 0$. In other words, the criterion for synchrony
becomes more stringent, it becomes more challenging to find
non-trivial clusters. Conversely, $T$ should be sufficiently large for
clusters to emerge. These observations are reflected in the results
shown in Fig. \ref{fig:results_bp}c. The size and performance of the
network decrease as a function of $\epsilon$ and increase as a
function of $T$. The original gradient descent method is the limit
case when $T \rightarrow \infty$ and $\epsilon \rightarrow 0$.

We find for the MNIST example that with $T=5$ (red curve) and $\epsilon=5 \times 10^{-2}$, the performance of the network trained with FB-GradDesc (accuracy = 95.73\%) remains within $\pm 2\%$ of the original network and gradient descent (97.67\%). However, with FB-GradDesc, the network could be compressed by a factor of 5 and 6 in the hidden layers, respectively.


\section{Discussion}\label{sec:Discussion}
We discussed three levels of graph symmetries: \textit{fibrations}, \textit{coverings}, and \textit{automorphisms}. Automorphisms are the traditional version of symmetries, where the graph structure is preserved. However, in real-world datasets, the data often lacks automorphism symmetries \cite{huang_approximately_2023}. On the other hand, building upon previous research \cite{boldi_fibrations_2002}, fibration (resp., covering) symmetries represent a natural extension of automorphism symmetries, as a consequence of local in-isomorphism (resp., local isomorphism) properties. As fibration and covering symmetries do not preserve the adjacency of graphs, they represent weaker symmetry levels than automorphism symmetries. Nevertheless, they serve as a powerful tool for analyzing the dynamics of networks by preserving the dynamics of nodes.

In Section \ref{sec:FibrationGNN}, we employ the formalism of symmetries to analyze the properties of GNNs, which are considered effective models for representation learning of graph-structured data \cite{zhou_graph_2020}. The effectiveness of such models serves as a clear example of how the symmetry of the domain (e.g., graphs) provides a powerful inductive bias (e.g., permutation invariance) \cite{bronstein_geometric_2021}. We demonstrate that GNNs are, at most, as powerful as the Fibration test in distinguishing graph structures (Theorem \ref{lemma}). Additionally, we establish conditions for aggregation and graph readout functions, under which the resulting GNN is as powerful as the Fibration test (Theorem \ref{theorem}). This result indicates that the expressive power of GNNs is indeed lower or equal to that proposed by Xu {\it et al.} \cite{xu_how_2019}.

The previous results imply that imposing stronger symmetry levels can enhance a model's expressive power. Zhang {\it et al.}
\cite{zhang_expressive_2023} mention that the limitation of GNNs in capturing global information limits its expressive power (Fig. \ref{fig:summary}b). To increase expressive power new GNNs have to satisfy the inductive bias induced by isomorphisms \cite{de_haan_natural_2020,thiede_autobahn_2022,purgal_improving_2020} (blue area in \ref{fig:summary}b). A model with high expressive power may potentially achieve high performance in certain tasks, but good performance is not guaranteed without proper training and correct configuration. On the flip side, relaxing the symmetry constraints induces a larger inductive bias, which may benefit learning if this bias is consistent with the learning task.

An empirical observation we have made here with MLPs is that the data and stochastic gradient descent (SGD) induce node synchronization in a network. By collapsing the corresponding nodes, we reduce the network size without affecting performance. This aligns with recent SGD theory \cite{chen_stochastic_2023}, which states that input and output trees of collapsed nodes are identical (i.e. covering symmetry), trapping the forward and backward calculation in an invariant set due to the permutation symmetry of MLPs. A similar form of ``neural collapse" has been observed in the last layers of deep networks trained on classification \cite{papyan_prevalence_2020}, but it is weaker and only within a class. The observation is that network weights converge to low-rank connectivity matrices. The corresponding symmetry is referred to as ``Simplex Equiangular Tight Frame" and differs from a covering symmetry.

Baker et al. \cite{baker_low-rank_2024} suggest this symmetry arises from low-rank gradients produced by backpropagation. While this inductive bias offers several advantages \cite{kothapalli_neural_2023}, it may not always be beneficial for specific learning tasks. The redundancy of nodes forms the basis of regularization techniques such as redundancy reduction \cite{cogswell_reducing_2016} or injecting noise with dropout \cite{srivastava_dropout_2014}, which break the symmetry induced by SGD to achieve a more expressive network without increasing parameters, thus reducing the need for larger datasets.

In summary, we have shown:
\begin{enumerate}
    \item Different graph symmetries affect the expressive power and inductive bias of models involving graph-structured data. In particular, we observed that local symmetries (e.g., fibrations) induce a strong inductive bias, which can be useful for improving the ability of machine learning models to generalize to unseen data. 
    \item Fibration symmetries provide a method for compressing graph-structured data, improving memory scalability and computations in graph neural networks (GNNs). This reflects the relationship between symmetry and complexity: a graph with more symmetries can be more compactly described and tends to have smaller Kolmogorov complexity values\footnote{Kolmogorov complexity of an object is the length of a shortest computer program that produces the object as output.} \cite{zenil_correlation_2014}.     
    \item The existence of fibrations in deep neural networks (DNNs) implies the synchronization of the activity of several nodes, which is useful for designing a new optimization method called Fibration-Gradient Descent. This new method allows for updating connection weights and building smaller networks without sacrificing performance.
\end{enumerate}

Our results show that applying local symmetries in deep neural networks is beneficial. Extending local symmetries to other domains is advantageous in Deep Geometric Learning. Local symmetries offer more flexible preservation of properties than global symmetries, making models more adaptable to local patterns. 


\section{Methods}\label{sec:Methods}

\subsection{Graph symmetries: Automorphisms, Coverings and Fibrations}\label{sec:preliminaries}

\begin{figure}[H]
\centering
\includegraphics[scale=0.115]{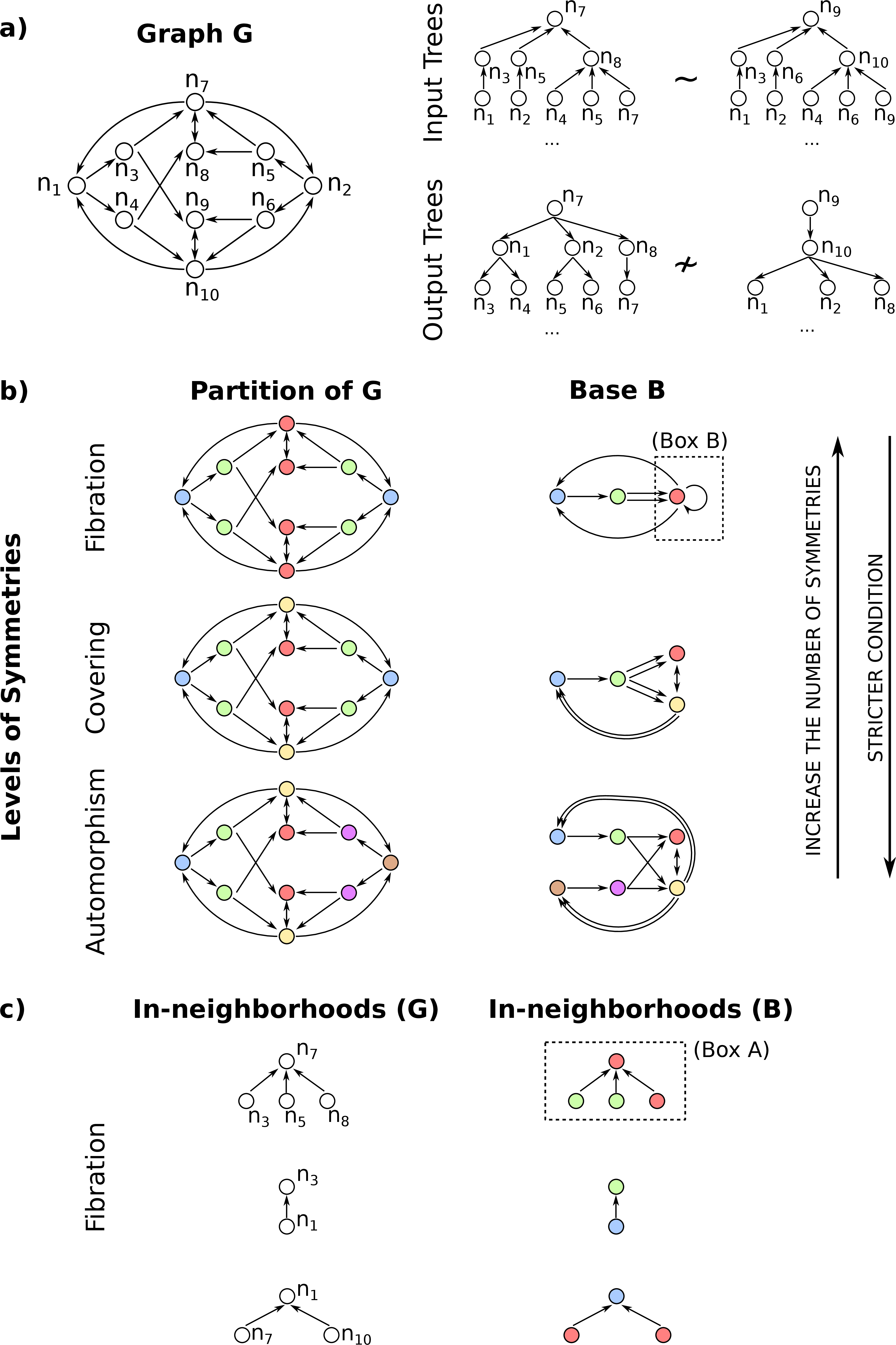}
\caption{\textbf{Symmetries of a graph}. \textbf{a)} In a graph $G$, each node has an associated input and output tree. Different nodes may have isomorphic trees (e.g. $T_{n_7} \sim T_{n_9}$) or not (e.g. $\hat{T}_{n_7} \nsim \hat{T}_{n_9}$). \textbf{b)} For each level of symmetry (i.e., fibrations, coverings, automorphisms), there is a partition of the nodes $N_G$ induced at each level. The graph $G$ can be compressed into a smaller graph $B$ called base. The nodes of the base $B$ represent equivalence classes (called fibers or balanced coloring), and the edges in the base respect the lifting property such that they preserve the information of the input trees of the nodes in $G$. For fibration (resp., covering) symmetries, nodes in $G$ within the same class (fibers) have isomorphic input trees (resp., input and output trees). In the automorphism symmetries, for nodes in $G$ within the same class (called orbits) there is an automorphism that maps one to another. \textbf{c)} By replacing the node color-IDs of $G$ in the in-neighborhoods with the class ID (e.g., $n_7 \mapsto $ red), we obtain the in-neighborhoods of the nodes in $B$.}
\label{fig:main_fig}
\end{figure}

In this work, we are interested in symmetries of directed graphs. A directed graph $G$ consists of a set $N_G$ of nodes and a set $A_G$ of edges\footnote{Unless otherwise specified, both $N_G$ and $A_G$ are finite.}. Each edge is associated with a source and a target node, and one way to represent the full structure of the graph is its incidence matrix, or its (possibly non-binary) adjacency matrix (as
long as you do not care about the identity of distinct parallel edges). The in-neighborhood (resp., out-neighborhood) of a node $u$ in $G$ is the set of nodes $N(-,u)$ (resp., $N(u,-)$) that have an edge
to (resp., from) the node $u$. Non-directed graphs are treated as directed graphs with a symmetric adjacency matrix, and $N(u) = N(-,u) = N(u,-)$. In Fig. \ref{fig:main_fig}a, we show a graph $G$ with 10
nodes, labeled as $N_G = \{n_1, n_2, ..., n_{10}\}$.

A graph homomorphism $\Phi: G \rightarrow H$ from a graph $G$ to a graph $H$ is a pair of functions $\phi_N: N_G \to N_H$ and $\phi_A: A_G \rightarrow A_H$ that map adjacent nodes of $G$ to adjacent nodes
in $H$. An \textit{automorphism} of $G$ is a homomorphism $\Phi_{auto}: G \to G$ whose components $\phi_{N}$ and $\phi_{A}$ are both one-to-one maps. In other words, an automorphism is a permutation
of nodes that preserve the adjacency. For example, in Fig. \ref{fig:main_fig}, the graph $G$ has two automorphisms: the trivial automorphism (i.e., the identity) and the vertical reflection, in cycle notation: $\Phi_{auto}=(n_1) (n_3, n_4) (n_7,n_{10}) (n_8, n_9) (n_5, n_6) (n_2)$.

For any graph $G$, the set of automorphisms form a symmetry group
(they satisfy associativity, have identity and inverse and are
composable). The symmetry group induces a class partition of $N_G$,
called the orbital partition, where two nodes $u,v \in N_G$ belong to
the same class if and only if there exists an automorphism $\Phi_{auto}$ such as $\phi_N(u)=v$. In Fig. \ref{fig:main_fig}b, we
show the partition of $G$ induced by the automorphisms, where each
color indicates a class (orbit) in the partition; e.g., the red class
is composed of nodes $n_8$ and $n_9$ because $n_8$ is the vertical
reflection of $n_9$.

Finding the automorphism symmetries of a graph can be
challenging for large graphs. The notion of input and output trees is a powerful tool for this task \cite{morone_fibration_2020} because they illustrate the flow within the graphs. For every node $u \in N_G$, there is a corresponding input tree $T_u$ (resp. output tree $\hat{T}_u$) that represents the set of all paths of $G$ ending in (resp. starting from) $u$. In Fig. \ref{fig:main_fig}a, we show the input and output trees corresponding to the $n_7$ and $n_9 \in
N_G$. Note that these trees are typically infinite even for finite
graphs, due to the presence of cycles.

An isomorphism between input trees $T_u$ and $T_v$ is defined as a bijective map
$\tau: T_u \rightarrow T_v$, which maps one-to-one the nodes and edges
of $T_u$ to nodes and edges of $T_v$. We use the notation $T_u \sim
T_v$ when there is an isomorphism between the input trees; otherwise,
we write $T_u \not\sim T_v$. The same definition and notation work for output trees.
For example, in Fig. \ref{fig:main_fig}a,
$T_7 \sim T_9$ but $\hat{T}_7 \not\sim \hat{T}_9$. An interesting
property is that if $\Phi$ is an automorphism of $G$, and $u, v \in
N_G$ such that $\phi_N(u) = v$; then $T_u \sim T_v$ and $\hat{T}_u
\sim \hat{T}_v$.  Using this property, we can identify nodes not in
the same class in the orbital partition induced by automorphisms. For
example, $n_7$ and $n_9$ of $G$ are in different orbital classes
(i.e., they have different colors) -- see row ``Automorphism" in
Fig. \ref{fig:main_fig}b.  Note that this condition is necessary but
\emph{not sufficient} for $\Phi_{auto}$ to be an automorphism.  That
is, there are nodes in the graph with $T_u \sim T_v$ but $\phi_N(u)
\ne v$ for any $\Phi_{auto}$ in the symmetry group of the graph.  This
extra symmetry is captured by the fibration.

Building upon this idea, we introduce the concept of \textit{fibration symmetry} \cite{boldi_fibrations_2002,morone_fibration_2020} (or minimal fibration\footnote{Boldi and Vigna \cite{boldi_fibrations_2002} present a more general definition of fibrations of graphs. In this context, \textit{minimal} refers to the minimal number of fibers \cite{gonzalez-aguilar_nonlinear_2015}.}). The \textit{fibration symmetry} of $G$ is a surjective homomorphism $\Phi_{fib}: G \to B$ such that 

\begin{equation}
    \label{eq:fibration}
    \forall u,v \in N_G: \phi_N(u)=\phi_N(v) \in N_B \iff T_u \sim T_v.
\end{equation}

In this definition, $B$ is called the \textit{fibration base}\footnote{The graph $B$ is unique up to isomorphism.} of $G$. We explain below how $B$ and $\phi_A$ can be constructed. Given $x
\in N_B$, the set $\phi_N^{-1}(x)$ (i.e. preimage of $U$) is called
the \textit{fiber} over $U$. Eq. (\ref{eq:fibration}) indicates that
two nodes are in the same fiber if and only if they have isomorphic
(indistinguishable up to relabeling of nodes) input trees.

Note that the definition of fibration symmetry considers only
the input trees of the nodes, not the output trees. We can consider a
stronger definition of symmetry. A \textit{covering symmetry} of a
graph $G$ is a surjective homomorphism $\Phi_{cov}: G \to B$ such that
\begin{equation}
    \forall u,v \in N_G: \phi_N(u) = \phi_N(v) \in N_B \iff T_u \sim T_v \quad \text{and} \quad \hat{T}_u \sim \hat{T}_v.
    \label{eq:covering}
\end{equation}
In this case, nodes in the same fiber have indistinguishable input and
output trees.

Just like automorphisms, the definitions of fibration and covering symmetries based on Eq. (\ref{eq:fibration}) and (\ref{eq:covering}) induce a partition of $N_G$, i.e. classes (fibers) of nodes that are equivalent because they have the isomorphic input trees (in case of fibrations) or isomorphic input and output trees (in case of coverings). In Fig. \ref{fig:main_fig}b, we show the partitions of a graph $G$ based on different symmetries (i.e., automorphisms, coverings, fibrations). All these partitions induced by the symmetries are \textit{balanced} \cite{gonzalez-aguilar_nonlinear_2015} since all nodes in the same fiber/orbit receive equal inputs from nodes of a given fiber/orbit. For this reason, the problem of finding symmetries is typically addressed using ``minimal balanced coloring" algorithms (see Section \ref{sec:met_fibration}).

It is important to note that the condition of automorphism symmetries in a graph $G$ is stricter than the condition of covering symmetries, and the latter is stricter than the
condition fibration symmetries. This is reflected in the number of symmetries, the number of partition classes, and the number of nodes per class. More precisely, the orbital partition induced by automorphisms is finer\footnote{A partition $P_1$ is finer than a partition $P_2$ if every fiber of $P_1$ is a subset of some fiber of $P_2$.} than or equal to the fiber partition induced by coverings, and the latter is finer than or equal to the fiber partition induced by fibrations (see Fig. \ref{fig:main_fig}b).

A \textit{global property} of a graph $G$ is a characteristic that
remains unchanged under isomorphisms of $G$, making it inherent to the
graph's structure. Examples include the size of the graph, connectedness, node degrees, etc. On the other hand, there
are other graph characteristics called \textit{local properties}, as
they remain invariant under local isomorphisms. This type of function
only requires that its restriction on neighborhoods $N(u)$ be an
isomorphism. For example, node degrees and input/output trees are
local properties. We are interested in the dynamics of the nodes as a property. It can be seen as a local and global property, depending on the context and how the dynamics are modeled. If the dynamics of a node $u$ depend on its neighborhood $N(u)$, then it can be considered a local property. On the other hand, the set of the states of all nodes represents the state of the entire graph, so it is considered a global property. 

All these considerations motivate us to distinguish three levels of
symmetries (see Level of Symmetries in Fig. \ref{fig:main_fig}b):
\begin{enumerate}
    \item \textbf{Automorphisms:} Nodes in the same class are entirely
      indistinguishable in terms of global (and local) properties.
    \item \textbf{Coverings:} Nodes in the same class have
      indistinguishable input and output trees, but may still be
      distinguishable in terms of global properties.
    \item \textbf{Fibrations:} Nodes in the same class have
      indistinguishable input trees, but may still have
      distinguishable output trees.
\end{enumerate}

Note that automorphism (resp., covering) symmetries are isomorphisms (resp., local isomorphisms) of a graph. Meanwhile, fibration symmetries are local
in-isomorphisms, i.e. homomorphisms such that their restriction on
in-neighborhoods is an isomorphism.

The automorphism symmetries are unique in preserving global
properties, whereas the other two levels of symmetries only preserve
certain local properties. This distinction between global and local
symmetries is also evident in other disciplines. For example, in
differential geometry, a global symmetry of a surface (or
diffeomorphism) implies that the number of holes or boundaries remains
constant. Therefore, a sphere (without holes) and a torus (with a
hole) are not equivalent at the level of diffeomorphisms. However, a
local diffeomorphism exists between them, so the surface appears flat
in small neighborhoods in both cases.

Another example, a bit more complex, arises in physics. In quantum
field theory, a global symmetry implies that fields undergo
transformations that remain constant across all space-time
\cite{dobado_global_1997}. An example is temporal invariance, which
implies the conservation of energy by Noether's theorem. On the other
hand, a local symmetry implies that fields can undergo local
transformations that vary in space and time, such as gauge
transformations. For instance, all four fundamental interactions (i.e. gravity, electromagnetism, weak interaction, and strong interaction) are described by gauge symmetries that maintain the Lagrangians invariant \cite{berghofer_gauge_2023}.

So far, we have not provided a complete definition of a base
$B$ for a fibration symmetry because Eq. (\ref{eq:fibration})
only gives us information about $N_B$ but not about the edges of
$B$. One way to define them is based on the structure of the
in-neighborhoods of the graph $G$. To do this, we follow the next
steps illustrated in Fig. \ref{fig:main_fig}c:

\begin{enumerate}
\item Calculate the partition of the nodes of $G$ (see colored nodes
  in Partition of $G$ - Fig. \ref{fig:main_fig}b).
\item Identify the in-neighborhood for all nodes of $G$ (see
  In-neighborhoods ($G$)).
\item Replace the node IDs in the in-neighborhoods with the class ID
  (i.e., color) to obtain the in-neighborhoods of the nodes in the
  base $B$ (see In-neighborhoods ($B$)).
\item Connect the nodes of $B$ using the information from the
  in-neighborhoods (see Boxes A and B in
  Fig. \ref{fig:main_fig}). Note that each node in the base $B$ can
  receive more than one edge from another.
\end{enumerate}

Using this procedure, the base $B$ can be a directed multigraph (i.e. there may be more than one edge with the same direction between two nodes) even if the graph $G$ is not. By convention, if $G$ is undirected, we consider each edge bidirectional. This method of constructing $B$ is analogous to the lifting property presented by Boldi et al. \cite{boldi_fibrations_2002}. The base of a graph can be interpreted as a way to summarize the presence of structural redundancies due to symmetries.

An immediate consequence of the definition of fibration symmetry is that each node $u$ in the original graph $G$ has in-neighborhood isomorphic to the in-neighborhood of the corresponding node $\phi(u)$ in the base $B$. Therefore, all nodes in the same fiber
have the same in-neighborhood. This is important for the analysis of dynamics on graphs where the state of each node depends solely on its in-neighbors (e.g., graph convolutional networks \cite{zhou_graph_2020}, gene regulatory networks \cite{barbuti_survey_2020}), because \textbf{the existence of a fibration $\phi: G \rightarrow B$ forces all nodes in the same fiber to remain always in the same state (referred to as ``node synchronization")}. This point is crucial for our work.

\subsection{Inductive bias induced by symmetries}\label{sec:InductiveBias}

A GNN is a mapping that takes a graph as input and produces a vector representation of the graph. In graph representation learning, it is expected that if two graphs are isomorphic (i.e., $G_1 \sim_{iso} G_2$), then their representation vectors are the same (i.e., $\text{GNN}(G_1) = \text{GNN}(G_2)$). This fact is known as the \textit{inductive bias induced by isomorphisms} or \textit{inductive bias induced by permutation invariance}. As shown in Fig. \ref{fig:summary}a, the equivalence relation $\sim_{iso}$ induces a partition in the Space of Graphs $\mathcal{X}$ (e.g. blue partition with 6 classes). The \textit{expressive power} of a GNN is defined as its ability to distinguish non-isomorphic graphs \cite{bronstein_geometric_2021} and is characterized by the number of isomorphism classes with different representation vectors. This definition of expressive power can be extended to any map $F: \mathcal{X} \rightarrow Y$, where $Y$ is a vector space. There exists a bound for the expressive power of any map $F$ with the inductive bias induced by isomorphisms, which is equal to the number of isomorphism classes (e.g. in Fig. \ref{fig:summary}b, $E_{F_{iso}} \leq 6$).

Usually, GNNs are interpreted as continuous and differentiable extensions of the well-known \textit{Weisfeiler-Lehman (WL) Isomorphism Test} \cite{weisfeiler_reduction_1968}. WL test is a method that, while it can indicate whether two graphs are not isomorphic, cannot provide conclusive evidence that they are isomorphic (see details in Supplementary Information). Both GNNs and the WL test iteratively aggregate information from local in-neighborhoods and use this information to update the features of each node. Xu {\it et al.} \cite{xu_how_2019} established that a GNN is at most as expressive as the Weisfeiler-Lehman (WL) Isomorphism Test (see the statements in Supplementary Information). In other words, these theorems ensure that the expressive power of the WL test is a bound for the expressive power of GNNs. One way to estimate this bound is as follows. The WL test takes two graphs $G_1$ and $G_2$ and computes its covering bases $B_1$ and $B_2$. If the bases $B_1$ and $B_2$ are different, the test will say that $G_1$ and $G_2$ are not isomorphic (see Supplementary Information). We define the relation $G_1 \sim_{cov} G_2$ to indicate that two graphs have the same covering bases, which induces a partition in the Space of Graphs $\mathcal{X}$ (see red partition with 5 classes in Fig. \ref{fig:summary}a). We say that a map $F$ has an \textit{inductive bias induced by coverings} when $F(G_1)=F(G_2)$ for any two graphs $G_1 \sim_{cov} G_2$ (e.g., WL test). For these maps, the maximum expressive power is the number of covering classes in the Space of Graphs (e.g., in Fig. \ref{fig:summary}b, $E_{F_{cov}} \leq 5$).

The calculation of the expressive power of the Fibration Test is analogous to the estimation we made for the WL test. We define the relation $G_1 \sim_{fib} G_2$ to indicate that two graphs have the same fibration bases, which induces a partition in the Space of Graphs $\mathcal{X}$ (see yellow partition with 4 classes in Fig. \ref{fig:summary}a). A map $F$ has an \textit{inductive bias induced by fibrations} when $F(G_1)=F(G_2)$ for any two graphs $G_1 \sim_{fib} G_2$ (e.g., Fibration test, GNNs). For these maps, the maximum expressive power is the number of fibration classes in the Space of Graphs (e.g., in Fig. \ref{fig:summary}b, $E_{F_{fib}} \leq 4$).

\subsection{Algorithm to find fibration symmetries}\label{sec:met_fibration}

The algorithm to find fibration symmetries of a graph $G$ is based on the ``minimal balanced coloring" algorithm \cite{gonzalez-aguilar_nonlinear_2015}. In particular, we used Kamei \& Cock algorithm \cite{kamei_computation_2013} to construct a minimal balanced coloring of a graph (see Alg. \ref{alg:fibr}), which is associated with the base $B$. It is worth noting that Alg. \ref{alg:fibr} is, in fact, a part of the WL test. The result of Alg. \ref{alg:fibr} allows us to construct the base and its colors match the i-colors in WL test (see an example and details in Table 1 of Supplementary Information). Code is available in \hyperlink{https://www.github.com/MakseLab}{Github}.

\begin{algorithm}
\caption{Fibration symmetries}\label{alg:fibr}
\begin{algorithmic}
\Require $G=(N_G,A_G)$
\State $\mathcal{L}^{(in)}_{0,i}(u) \gets d^{(in)}(u), \quad \forall u \in N_G$.
\State $L^{(in)}_{0,i} \gets \{ (u,\mathcal{L}^{(in)}_{0,i}(u)), u \in N_G \}$.
\Repeat
\State $L^{(in)}_{t,i} \gets F(G_i,L^{(in)}_{t-1,i})$
\Until $L^{(in)}_{t,i} = L^{(in)}_{t-1,i}$
\Ensure $L^{(in)}_{t,i}$
\end{algorithmic}
\end{algorithm}


\subsection{Deep Neural Networks}
The design of a suitable architecture for a DNN heavily depends on the task at hand. In the following sections, we describe two types of architectures that we employ for the task of image classification and graph regression. 

We utilize the notations $\mathcal{D} = \{(x^{(n)},y^{(n)})\}^{N}_{n=1}$ and \textit{Loss} to indicate the dataset and loss function employed in the training and evaluation stages. Each element of $D$ is a tuple of an object $x^{(n)}$ (e.g. image or graph) and its label $y^{(n)}$ (e.g. classes of the image or properties of the molecule). DNN generates a prediction $\hat{y}_n$ for each $x^{(n)}$, which is compared to the label $y^{(n)}$ using \textit{Loss}.

\subsubsection{Multilayer perceptrons}
One of the most popular architectures of DNNs is the multilayer perceptron (MLP). The nodes of the network are organized into multiple layers $l$: an input layer ($l=1$), one or more hidden layers ($l=2,3,...,M-1$), and an output layer ($l=M$). Each layer $l \in \{1,2...,M\}$ consists of $d_l$ neurons, and connections between neurons exist between layers $w^{l-1 \rightarrow l}_{i \rightarrow j}$. Each unit in a layer receives a total input $I_i^l \in \mathbb{R}$, applies specific mathematical operations (e.g., a nonlinear function $F$) to this input, and generates an activity $h_i^l \in \mathbb{R} $. Mathematically, 
\begin{align} 
h^l_i &= F(I^l_i) \label{eq:MLP_act}, \\
I^l_i &= \sum_{j \in l-1} w^{l-1 \rightarrow l}_{j\rightarrow i} h^{l-1}_j + x_i \delta_{l,1}
\label{eq:MLP_inp}
\end{align}
where $x \in \mathcal{D}$ is the input in the first layer ($l=1$), and $w^{l-1 \rightarrow l}_{j \rightarrow i}$ indicates the weight of the connection $j \in l-1 \rightarrow i \in l$, respectively. We use the symbol $\delta$ to represent the Kronecker delta. Note that, in this case, $x^{(n)} \in \mathbb{R}^{d_1}$ is a vector. 

We emphasize that in reality, $h_i = h_i(x)$ indicates the activity of node $i$ when the input of the network is $x \in \mathcal{D}$. However, we simplify the notation whenever it does not lead to confusion. Additionally, we define
\begin{equation}
h_i(\mathcal{D}) = \frac{1}{|\mathcal{D}|} \sum_{x \in \mathcal{D}} h_i(x).
\end{equation} 

\subsubsection{Graph neural networks}
Graph Neural Networks (GNNs) are a class of deep neural networks designed to process data represented as graphs (i.e. $x^{(n)}$ is a graph) for node-level, edge-level, and graph-level prediction tasks. 

In the inference stage, each input graph $G \in \mathcal{D}$ is represented with an adjacency matrix $\mathbf{A} \in \mathbb{R}^{|N_G| \times |N_G|}$, a node representation matrix $H^{(1)} \in \mathbb{R}^{|N_G| \times d_1}$, and edge properties $E \in \mathbb{R}^{|A_G| \times f}$. Sometimes, GNN operations use the matrix $\hat{\mathbf{A}} = \mathbf{A} + \mathbf{I}$ and diagonal degree matrix $\hat{\mathbf{D}}_{ij} = \delta_{ij} \sum_k \hat{\mathbf{A}}_{kj}$ in their computation. In this work, we focus on four types of operations: a) graph convolution (GC), b) graph edge-conditioned convolution (GEC) c) graph isomorphism (GI), and d) graph attention (GA), which are defined by the following equations

\begin{align} 
\textbf{GC}: H^{(l+1)} &= F(\hat{\mathbf{D}}^{-\frac{1}{2}} \hat{\mathbf{A}}^T \hat{\mathbf{D}}^{-\frac{1}{2}} H^{(l)} W^{(l)}) \label{eq:GC} \\ 
\textbf{GEC}: h^{(l+1)}_i &= F(h^{(l)}_i \Theta^{(l)} + \sum_j A_{ji} h^{(l)}_j W^{(l)}(e_{ji})) \label{eq:GEC} \\
\textbf{GI}: H^{(l+1)} &= F^{(l)}(((1+\epsilon^{(l)})\mathbf{I} + \hat{\mathbf{A}}^T) H^{(l)}) \label{eq:GI} \\
\textbf{GA}: H^{(l+1)} &= F\left(\alpha^{(l)} H^{(l)} W^{(l)} \right) \label{eq:GA}
\end{align}

where $H^{(l)} \in \mathbb{R}^{N \times d_l}$ is the matrix of node representation in the layer $l=1,...,L$ ($h^{(l)}_i$ is ith-row of this matrix). For GC and GA, $W^{(l)} \in \mathbb{R}^{d_l \times d_{l+1}}$ is a matrix of trainable parameters; while for GEC is a function $W^{(l)}: \mathbb{R}^{f} \rightarrow \mathbb{R}^{d_l \times d_{l+1}}$ (with trainable parameters) that maps edge properties $e_{ij}$ (row of $E$) into a matrix. $F$ represents a nonlinear activation, such as the ReLU or sigmoid function (for GC, GCE, and GA), or something more complex like an MLP (for GI). For the GI operation, the parameter $\epsilon^{(l)}$ controls how the importance of the node's feature is weighted compared to the features of its neighbors, and generally takes values between -1 and 1. For GEC, the matrix $\Theta^{(l)} \in \mathbb{R}^{d_l \times d_{l+1}}$ is a matrix of trainable parameters. For GA, the attention matrix $\alpha^{(l)} \in \mathbb{R}^{|N_G| \times |N_G|}$ is computed following
\begin{align*}
    \alpha^{(l)} &= \text{softmax}(e^{(l)}) \\
    e^{(l)} &= \rho(H^{(l)}W^{(l)} b^{(l)}_1 \mathbf{1} + (H^{(l)}W^{(l)} b^{(l)}_2 \mathbf{1})^T) + m
\end{align*}
where $m_{ij} = \log(\mathbf{A}_{ji})$, $\rho$ is a nonlinear function (e.g. ReLU), $\mathbf{1}=[1,...,1] \in \mathbb{R}^{1 \times N}$, and $b^{(l)}_1, b^{(l)}_2 \in \mathbb{R}^{d_l} $ are vectors of trainable parameters.

\textbf{One important point to emphasize about GNN operations is that $h_i^{(l+1)}$ depends exclusively on the features of the in-neighborhood of node $i$ in layer $l$.}


\subsection{Datasets}\label{sec:datasets}

In Section \ref{sec:Performance}, we implemented GNNs for two graph regression tasks. The first task is the calculation of chemical properties in organic molecules. For this, we used samples from \hyperlink{https://pytorch-geometric.readthedocs.io/en/latest/generated/torch_geometric.datasets.QM9.html}{QM9} dataset, which is a publicly available repository containing information about the chemical and structural characteristics of small organic molecules. This dataset contains data for approximately 133,885 distinct organic molecules, including details such as total electronic energy, bond energy, atomization energy, dipole moment, and isotropic polarizability. Additionally, for each molecule in the dataset, comprehensive information about its molecular structure is provided, including three-dimensional atomic coordinates and atom types. This information is derived from ab initio quantum chemistry calculations utilizing the Quantum ESPRESSO program \cite{ramakrishnan_quantum_2014}.

The second task is the prediction of the overall survival time (in months) of patients with breast cancer. We used \hyperlink{https://pytorch-geometric.readthedocs.io/en/latest/generated/torch_geometric.datasets.BrcaTcga.html}{BRCA TCGA Pan-Cancer Atlas} dataset, which consists of 1,082 samples (i.e. patients). Each sample contains the same directed graph $G$ that represents biological interactions ($|A_G| = 271,771$) between genes ($|N_G| = 9,288$) \cite{rodchenkov_pathway_2020}. The difference between samples is the gene expression data (i.e., features of the nodes), which is represented by a real number per node \cite{gao_integrative_2013}.

For the work with  MLPs, we used classic image classification tasks. To accomplish this, we leveraged four publicly accessible datasets: MNIST \cite{deng_mnist_2012}, KMNIST \cite{clanuwat_deep_2018}, and FASHION \cite{xiao_fashion-mnist_2017}. Each of these datasets consists of 70k 28x28 grayscale images, with 60k designated for training and 10,000 for evaluation, as detailed below. As described below, all the datasets categorize the images into 10 classes
\begin{enumerate}
    \item \hyperlink{http://yann.lecun.com/exdb/mnist/}{MNIST} is a popular dataset consisting of images of handwritten digits from 0 to 9. 
    \item \hyperlink{http://codh.rois.ac.jp/kmnist/index.html.en}{Kuzushiji-MNIST}(KMNIST) is a dataset specifically created for recognizing handwritten Japanese characters known as Kuzushiji.
    \item \hyperlink{https://github.com/zalandoresearch/fashion-mnist}{FASHION} is a dataset with fashion-related images (e.g., t-shirts, trousers, dresses).
\end{enumerate}


\subsection{Fibration symmetries in GNNs}\label{sec:Fibrations_GNNs}

Proposition 1 of \cite{jegelka_theory_2022} establishes that if in a GNN, two nodes have the same trees then their states will be synchronized. This result is well-known in the theory of fibrations and groupoids \cite{deville2015modular}. In terms of fibration symmetries, this proposition can be stated as follows:

\begin{theorem}
Let $G$ be a graph. If two nodes $u, v \in N_G$ are in the same fiber, then node feature vectors $h_u^{(k)}$ and $h_v^{(k)}$ in a GNN are synchronized, i.e. $h_u^{(k)}$ = $h_v^{(k)}$.
\label{theorem3}    
\end{theorem}

\medskip

Using this proposition, we can reduce the number of operations in a GNN based on node synchronization. Since the number of operations of a GNN is derived from the graphs in the dataset; if we find the fibration symmetries within the dataset samples, it is possible to reduce the the number of operations. 
  
Based on these theoretical results, we propose two different ways to
train a GNN: standard and reduced form (see
Fig. \ref{fig:performance}). For the standard form, the
inputs to the GNNs are the samples $G$ from the graph-structured
dataset $\mathcal{D}$ without any prior processing; while for the reduced form, two steps are performed
\begin{enumerate}
    \item Search for the fibration symmetries ($\phi$) and bases ($B$) of the dataset samples ($G$).
    \item Train the GNN such that the inputs are the bases ($B$).
\end{enumerate}

To achieve this, we have to derive equations for operations (GC: graph convolution, GEC: graph edge-conditioned convolution, GI: graph isomorphism, and GA: graph attention) on the base graph $B$ that are compatible (i.e. preserving the information) with the operations on the original graph $G$. In Supplementary Information, we present the derivation of the equation for the GEC, GC, GA, and GI operation on the base graph $B$. 


\subsection{Fibration symmetries in MLPs}\label{sec:met_fibration_mlp}

We train MLPs for standard image classification tasks on three independent datasets $\mathcal{D}$: MNIST, KMNIST, and Fashion (see Section \ref{sec:datasets}). In all cases, the networks consist of two hidden layers (see Fig. \ref{fig:MLP}a). The 28x28 images $x \in \mathcal{D}$ are flattened into a 784-dimensional input vector. The weights are initialized randomly and updated using stochastic gradient descent to minimize the cross-entropy loss. 

It is important to note that detecting fibration symmetries through balanced coloring in graphs with continuous weight connections (e.g., MLP) can be complicated. However, if two nodes $i$ and $i'$ have the same activity $h_i$ and $h_{i'}$ for all inputs $x \in \mathcal{D}$, it is expected that they have isomorphic input trees, and therefore should belong to the same fiber. We observed that as the training progresses, clusters of synchronized nodes (or fibers) tend to emerge in the network. To formalize this idea, given two nodes $i$ and $i'$ in a layer $l$, we define
\begin{align*} 
d_{i,i'}(x) &= h^l_i(x) - h^l_{i'}(x) ,\\
\xi_{i,i'} &= \vert \frac{1}{|\mathcal{S}|} \sum_{x\in \mathcal{S}} d_{i,i'}(x)\vert
\qquad x \in \mathcal{S} \subset \mathcal{D}.
\end{align*}

We say two nodes are synchronized if and only if their distance is smaller than some bound, $\xi_{i,i'} < \epsilon_l$, across a subset of the full dataset, $\mathcal{S} \subset \mathcal{D}$. We select a distance tolerance of $\epsilon_l = 0.1$. Also, we compute the distance matrix between nodes $\Lambda_{i,i'} = \frac{1}{|\mathcal{S}|} \sum_{x\in \mathcal{S}} |d_{i,i'}(x)|$. 

\subsection{Fibration Gradient Descent}\label{sec:methods_mlp}

We implemented a new training algorithm for MLPs (represented by graphs $G$) using information related to its bases $B$. To achieve this, we found the equation of the dynamics in the base $B$ (see Supplementary Information).

For the original MLP $G$, at the outset of the learning process, connection weights are initialized randomly, leading to inherent variability in node activities within the network. During this stage, the likelihood of synchronized nodes is minimal (i.e., non-trivial fibers are absent). Toward the process's end, the weights are adjusted to optimize the loss function, potentially resulting in certain nodes becoming synchronized (i.e., fibers emerge). This observation suggests that the learning process applied to $G$ gives rise to the emergence of fibers. In the algorithm proposed (Alg. \ref{alg:training_fibrations}), we use these fibrations to reduce the dimensionality of the original network. Additionally, we incorporate a modified equation into the learning process for the new network, ensuring compatibility with the training of the original network (see proof in Supplementary Information). The compatibility between the training equations enable us to achieve similar performances for networks of different sizes (where the base network is significantly smaller than the original network). Code is available in \hyperlink{https://www.github.com/MakseLab}{Github}.

\begin{algorithm}
\caption{Training using fibrations}\label{alg:training_fibrations}
\begin{algorithmic}
\Require Network $G=\mathcal{N}_0$, Dataset $\mathcal{D}$
\Ensure Network $B=\mathcal{N}_T$ (post-training)
\State $\mathcal{N} \gets \mathcal{N}_0$
\State $N \gets n$
\For {$t \in [0,T]$}
\State $\mathcal{N} \gets BP(\mathcal{N},\mathcal{D})$ \textbf{(training of $\mathcal{N}$)}
\State $\mathcal{P} \gets Fibrations(\mathcal{N},\mathcal{D})$ \textbf{(get fibrations of $\mathcal{N}$)}
\State $M \gets Partition(\mathcal{P})$ \textbf{(get transformation to new structure)}
\State $\mathcal{N} \gets Base(\mathcal{N},M)$
\EndFor
\end{algorithmic}
\end{algorithm}

Note that in Alg. \ref{alg:training_fibrations}, the size of the network changes during epochs, unlike the typical training process. The calculation of fibrations during training facilitates the reduction of the number of nodes. Details regarding the theoretical deduction of the algorithm are shown in Supplementary Information.

\backmatter
\bibliography{References_Main}

\end{document}